\documentclass[letterpaper, 10 pt, conference]{ieeeconf}  

\IEEEoverridecommandlockouts                              

\usepackage{graphics} 
\usepackage{epsfig} 
\usepackage{mathptmx} 
\usepackage{times} 
\usepackage{amsmath} 
\usepackage{amssymb,bm}  
\usepackage{graphicx}
\usepackage{color}
\usepackage{lipsum}

\usepackage[style=ieee,dashed=false]{biblatex}

\DeclareSourcemap{
  \maps{
    \map{
      \pertype{article}
      \step[fieldset=language, null]
      \step[fieldset=url, null]
      \step[fieldset=doi, null]
      \step[fieldset=issn, null]
      \step[fieldset=isbn, null]
      \step[fieldset=note, null]
      \step[fieldset=editor, null]
      \step[fieldset=urldate, null]
      \step[fieldset=file, null]
    }
  }
}
\DeclareSourcemap{
  \maps{
    \map{
      \pertype{inproceedings}
      \step[fieldset=language, null]
      \step[fieldset=url, null]
      \step[fieldset=doi, null]
      \step[fieldset=issn, null]
      \step[fieldset=isbn, null]
      \step[fieldset=note, null]
      \step[fieldset=editor, null]
      \step[fieldset=urldate, null]
      \step[fieldset=file, null]
    }
  }
}
\DeclareSourcemap{
  \maps{
    \map{
      \pertype{incollection}
      \step[fieldset=language, null]
      \step[fieldset=url, null]
      \step[fieldset=doi, null]
      \step[fieldset=issn, null]
      \step[fieldset=isbn, null]
      \step[fieldset=note, null]
      \step[fieldset=editor, null]
      \step[fieldset=urldate, null]
      \step[fieldset=file, null]
    }
  }
}

\title{\LARGE \bf
Estimation of Aerodynamics Forces in Dynamic Morphing Wing Flight
}

\author{Bibek Gupta$^{1}$, Mintae Kim$^{2}$, Albert Park $^{1}$, Eric Sihite$^{3}$, Koushil Sreenath$^{2}$ and Alireza Ramezani$^{1*}$
\thanks{$^{1}$Authors are with the Silicon Synapse Lab., Department of
Electrical and Computer Engineering, Northeastern University, Boston,
USA. Emails: gupta.bi, al.park, a.ramezani@northeastern.edu}%
\thanks{$^{2}$Authors are with the Department of Mechanical Engineering,
Univ. of California, Berkeley CA 94720,
USA. Emails: mintae.kim, koushils@berkeley.edu
}%
\thanks{$^{3}$Author is with the Department of Aerospace Engineering,
California Institute of Technology, Pasadena, USA. Email:
esihite@caltech.edu}%
\thanks{*Corresponding author.}
}

\begin{document}

\maketitle
\thispagestyle{empty}
\pagestyle{empty}

\begin{abstract}

Accurate estimation of aerodynamic forces is essential for advancing the control, modeling, and design of flapping-wing aerial robots with dynamic morphing capabilities. In this paper, we investigate two distinct methodologies for force estimation on \textit{Aerobat}, a bio-inspired flapping-wing platform designed to emulate the inertial and aerodynamic behaviors observed in bat flight. Our goal is to quantify aerodynamic force contributions during tethered flight, a crucial step toward closed-loop flight control. The first method is a physics-based observer derived from Hamiltonian mechanics that leverages the concept of conjugate momentum to infer external aerodynamic forces acting on the robot. This observer builds on the system's reduced-order dynamic model and utilizes real-time sensor data to estimate forces without requiring training data. The second method employs a neural network-based regression model, specifically a multi-layer perceptron (MLP), to learn a mapping from joint kinematics, flapping frequency, and environmental parameters to aerodynamic force outputs. We evaluate both estimators using a 6-axis load cell in a high-frequency data acquisition setup that enables fine-grained force measurements during periodic wingbeats. The conjugate momentum observer and the regression model demonstrate strong agreement across three force components ($F_x$, $F_y$, $F_z$). 


\end{abstract}

\section{Introduction}

Actively functional organs with inertial dominance are employed by animals to reorient their bodies and respond quickly to environmental stimuli. Airborne locusts, for example, utilize abdominal motion to regulate body orientation midair \cite{bomphrey_tomographic_2012,jusufi_active_2008,jusufi_righting_2010}. Similarly, vertebrate animals like lizards employ their intersegmental tail to adjust their body orientation \cite{libby_tail-assisted_2012}. Pronounced movements of such inertial appendages, like a tail, have known angular momentum implications, leading to significant changes in body momentum.

In contrast to these terrestrial examples, vertebrate flyers such as bats exhibit a complex interplay between inertial dynamics and aerodynamic forces \cite{riskin_bats_2009,thollesson_moments_1991,iriarte-diaz_whole-body_2011}. For instance, bats utilize inertial dynamics to perform maneuvers like heel-above-head landings. Even when high-angle-of-attack maneuvers cause aerodynamic forces to diminish, bats can still control their descent. Unlike insects, whose lightweight wings contribute minimally to inertial effects \cite{deng_flapping_2006-1}, bats have wings that constitute a significant portion of their body weight, amplifying the role of inertial forces.

\begin{figure}
    \centering
    \includegraphics[width=1\linewidth]{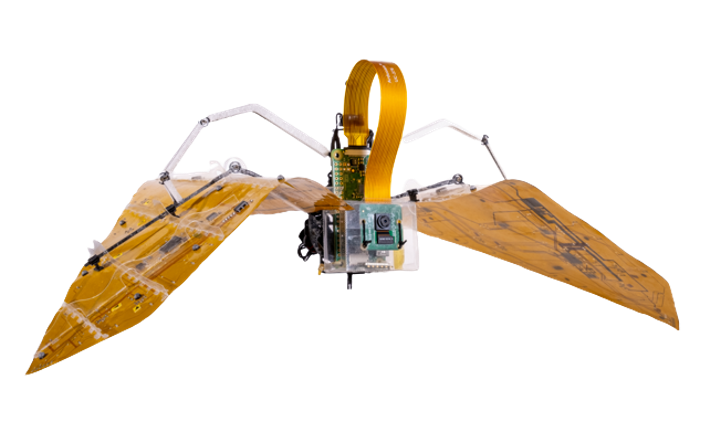}
    \caption{Shows Aerobat platform \cite{sihite_actuation_2023}. The platform is designed to study inertial and aerodynamic dynamics contribution roles in dynamic morphing wing flight.}
    \label{fig:cover-image}
\end{figure}

To investigate the role of dynamic morphing wings and bat-like flight characteristics, we designed and developed Aerobat \cite{sihite_wake-based_2022,sihite_computational_2020,ramezani_bat_2016,ramezani_biomimetic_2017,sihite_morphology-centered_2025}, shown in Fig.~\ref{fig:cover-image}. Aerobat features large morphing appendages (each approximately 0.4 grams) that flex and expand dynamically during each gait cycle (100 ms), contributing substantially to inertial dynamics and influencing body motion. The presence of these inertial forces complicates the estimation of aerodynamic forces, which are typically modeled in insect flight without concern for inertial contributions.

The primary objective of this work is to estimate the aerodynamic forces acting on Aerobat during hovering, which is critical for controlling the vehicle's flight \cite{dhole_hovering_2023,gupta_banking_2024,gupta_bounding_2024}. We explore two approaches: 1) conjugate momentum observer, and 2) a neural network-based regression model implemented as a multi-layer perceptron (MLP). We test these methods using a testbed that is based on a 6-axis load cell, which measures aerodynamic force contributions in a tethered configuration. The data obtained from this setup serves as ground truth to validate the performance of our estimators.


Theoretical work on momentum observers \cite{vorndamme_collision_2017,albuschaffer_dlr_2007} is extensive, particularly in legged locomotion, where these methods are widely used to monitor external forces acting on a system \cite{pitroda_conjugate_2024,krishnamurthy_optimization_2024}. However, their application in flapping wing flight remains largely unexplored, likely due to the difficulty in designing flapping robots with dynamic morphing wings, making the use of such observers seem unnecessary.



Multi-Layer Perceptron (MLP) is a class of neural networks designed to approximate complex and nonlinear functions \cite{hornik1989multilayer}. An MLP with at least one hidden layer and a sufficient number of neurons can act as a universal approximator, capable of approximating any continuous function given sufficient data to capture the underlying function that generates the observations \cite{cybenko1989approximation}. MLPs are well-suited for tasks such as force estimation in flapping-wing UAVs, where aerodynamic forces follow highly nonlinear relationships and depend on multiple variables, including actuation and experimental parameters. The specifics of the MLP-based force regression method are discussed in Section \ref{sec:estim}.

The key contributions of this work are twofold. First, we adopt two estimation methods: one based on classical Hamiltonian mechanics and another on a regression approach, to estimate forces in dynamic morphing wing flight, known for its complex wing-air interactions. Second, we experimentally validate these methods using tethered flapping datasets. 

 This work is organized as follows. In Section \ref{sec:hdw}, we describe the hardware specifications of the Aerobat platform and the force-measurement testbed. Section \ref{sec:model} presents the modeling approach for the Aerobat system. Section \ref{sec:estim} details the force estimation methods, including the conjugate momentum observer and regression techniques. Section \ref{sec:res} presents the experimental results and compares the performance of different estimation methods.

\section{Hardware Specifications}
\label{sec:hdw}


In this paper, we used two hardware components: 1) 
Aerobat platform (Fig.~\ref{fig:cover-image})  and 2) force measurement testbed (Fig.~\ref{fig:loadcell}). In the following, we briefly describe each hardware setup and its specifications.

\subsection{Aerobat}
The Aerobat possesses articulated wings that dynamically reconfigure during flight, mimicking bat morphology. With a 30 cm wingspan and 30g weight plus 20g payload capacity, its key feature is the ability to fold and expand wings during upstroke and downstroke. This morphing, driven by a "computational structure" of linkages and a single brushless DC motor, enables complex flapping gaits, potentially enhancing aerodynamic efficiency in bio-inspired flight \cite{lessieur_mechanical_2021}.


\subsection{Force Measurement Testbed}
\begin{figure}
    \centering
    \includegraphics[width=1\linewidth]{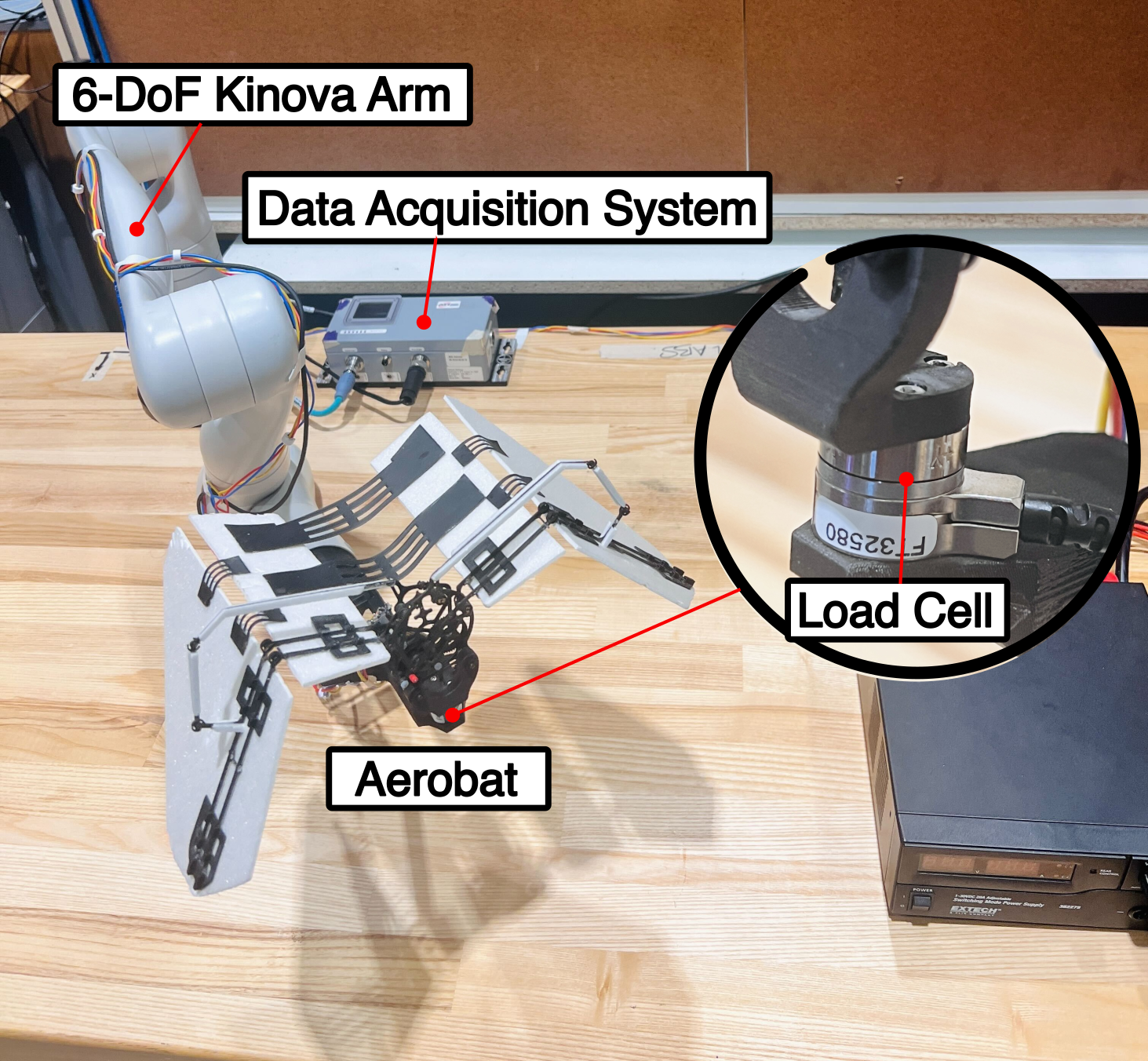}
    \caption{Measurement setup showing \textit{Aerobat} mounted on the end-effector of a Kinova robotic arm. A six-axis load cell is integrated at the interface to measure interaction forces and moments.}
    \label{fig:loadcell}
\end{figure}

The force measurement setup, shown in Fig.~\ref{fig:loadcell}, employs a Gen-3 6-DOF Kinova robotic arm to precisely position the Aerobat robot. A custom 3D-printed mount attaches the Aerobat to the arm's end-effector, incorporating an ATI Nano17 6-axis load cell. This transducer is capable of measuring forces up to ±16 N in the X and Y directions, ±28.2 N in the Z direction, and torques up to ±0.1 N·m. It connects to a NetBox for data acquisition via Ethernet. An axial fan, positioned 2 m in front of the Aerobat, was used to generate airflow toward its frontal plane. Wind speed was measured 5 cm upstream of the Aerobat using an anemometer. During stationary and tethered experiments, force and moment data were collected at a sampling frequency of 7 kHz.

\section{Modeling}
\label{sec:model}

\begin{figure}
    \centering
    \includegraphics[width=1\linewidth]{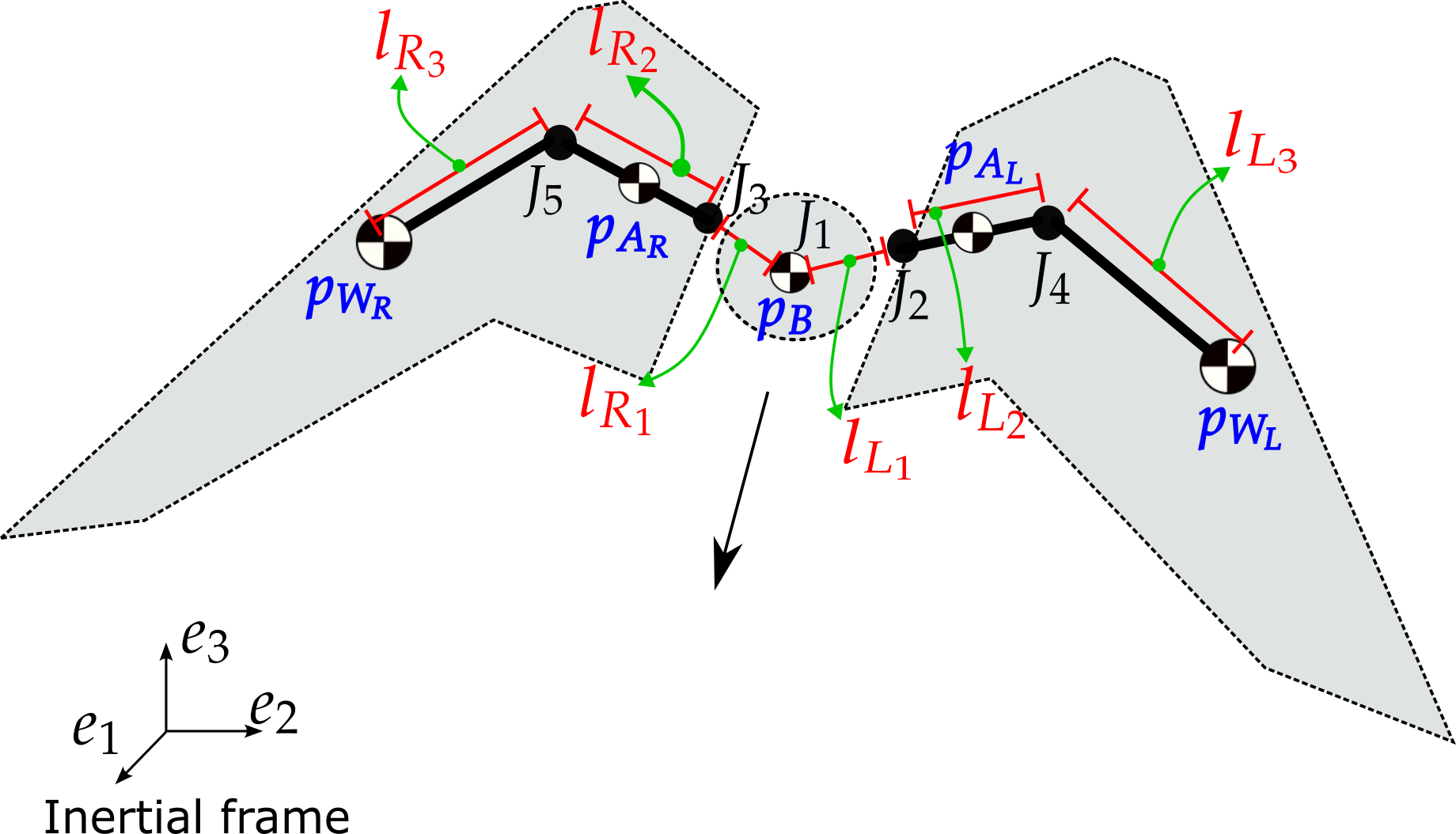}
    \caption{Aerobat kinematics consisting of five rigid bodies connected at joints $J_i$, $i = \{1,\dots,5\}$, which define the revolute joints. The vectors $\bm{p}_{B}$, $\bm{p}_{A
    _R}$, and $\bm{p}_{A_L}$ denote the body, right and left arm positions, while $\bm{p}_{W_R}$ and $\bm{p}_{W_L}$ represent the positions of the right and left wing masses. Length terms $\bm{l}_{R_j}$ and $\bm{l}_{L_j}$, $j = \{1,2,3\}$, describe the link dimension of the right and left wings, respectively. The configuration is parameterized relative to the inertial frame $\{ \bm{e}_1, \bm{e}_2, \bm{e}_3 \}$.}

    \label{fig:aerobat_model}
    \vspace{-0.05in}
\end{figure}

The Aerobat system, depicted in Fig.~\ref{fig:cover-image}, can be modeled as five rotating bodies connected by joints or hinges, as shown in Fig.~\ref{fig:aerobat_model}.

\begin{figure}
    \vspace{0.05in}
    \centering
    \includegraphics[width=0.8\linewidth]{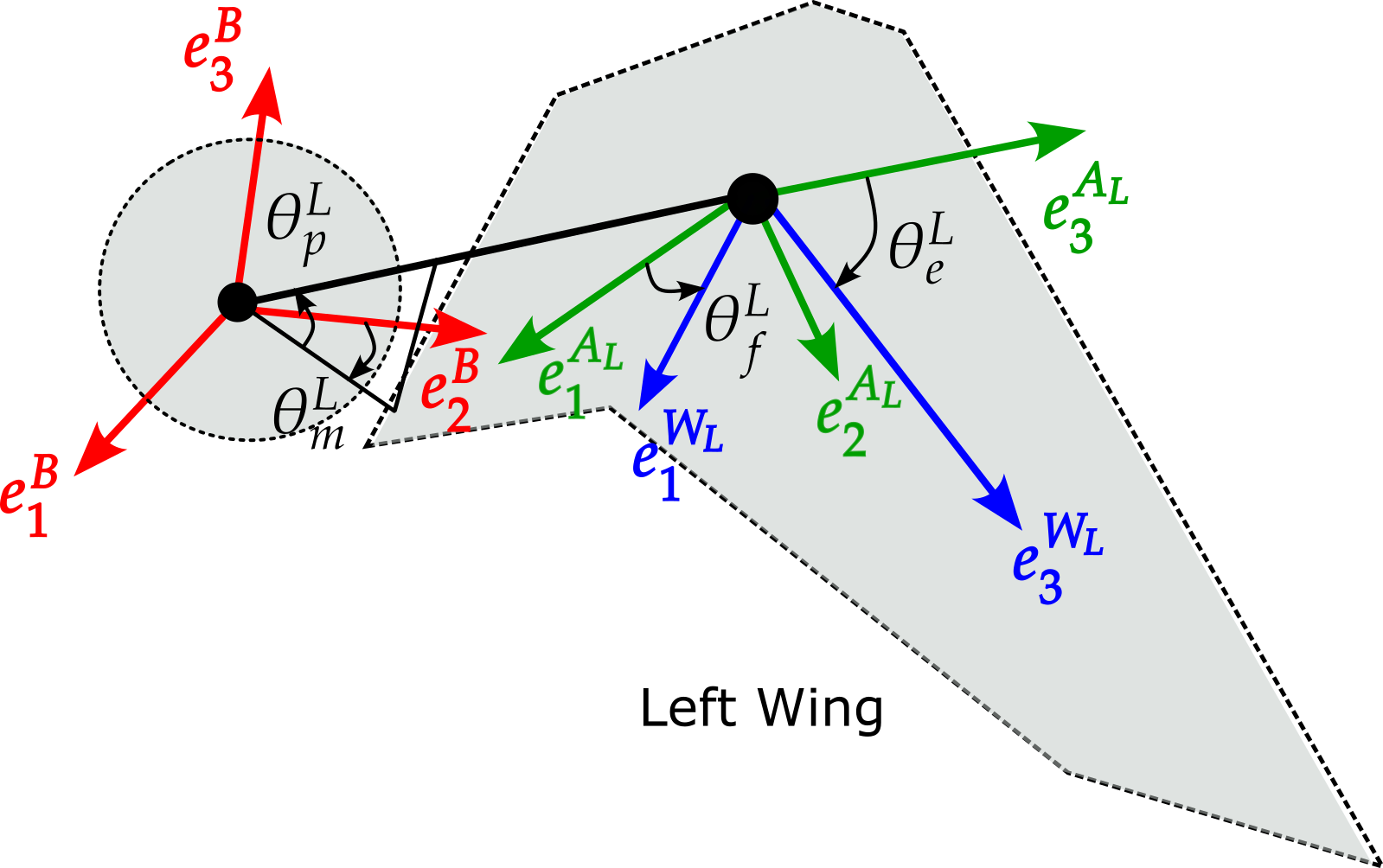}
    \caption{Aerobat biologically meaningful degrees-of-freedom angles: armwing plunging ($\theta_p$), mediolateral ($\theta_m$), elbow extension ($\theta_e$), and feathering ($\theta_f$). $\theta_p$ and $\theta_m$ are defined with respect to the body frame while $\theta_e$ and $\theta_f$ are defined with respect to the arm frame. The superscript $L$ and $R$ are used to represent left and right wing angles respectively.}
    \label{fig:aerobat_dof}
    \vspace{-0.05in}
\end{figure}

This armwing mechanism is designed to replicate the biologically relevant degrees of freedom (DoF) of a bat's armwing, as illustrated in Fig.~\ref{fig:aerobat_dof}. These DoFs are reflected in Aerobat's flapping wing flight:

\begin{itemize}
    \item The shoulder joint plunge angle $(\theta_p)$ controls the upstroke and downstroke of the wing, which forms the core flapping motion.
    \item The elbow extension/flexion angle $(\theta_e)$ expands the wing during the downstroke and retracts it during the upstroke, enhancing efficiency by reducing the negative lift during the upstroke.
\end{itemize}

In addition to these wing movements, we also consider the mediolateral motion $(\theta_m)$, which extends the wings forward, and the feathering motion $(\theta_f)$, which rotates the plane of the wing surface relative to the arm, affecting the angle of attack. 

Let a vector with a superscript notation represent the vector defined in a non-inertial coordinate frame and the vector without superscript is defined about the inertial frame, e.g. $\bm{x}^{B}$ is the vector $\bm{x}$ about frame $B$. The frame of references is illustrated in Fig. \ref{fig:aerobat_model} and the coordinate frame rotation of the five bodies can be defined as follows:
\begin{equation}
    \begin{aligned}
        \bm{x} &= R_B\,\bm{x}^{B}, \quad &
        \bm{x}^B &= R_{A_L}\,\bm{x}^{A_L} = R_{A_R}\,\bm{x}^{A_R} \\
        \bm{x}^{A_L} &= R_{W_L}\,\bm{x}^{W_L}, \quad &
        \bm{x}^{A_R} &= R_{W_R}\,\bm{x}^{W_R},
    \end{aligned}
\label{eq:rotation_matrices}
\end{equation}
where $R_B$ is the body rotation matrix in relation to the inertial frame, $R_{A_L}$ and $R_{A_R}$ are the rotation matrix of the left and right arms respectively about the body frame, and $R_{W_L}$ and $R_{W_R}$ are the rotation matrix of the left and right wings, respectively about their respective arm. 

The corresponding angular velocities for these rotation matrices must also be represented in the appropriate coordinate frames. Then we have the following rotation matrix and angular velocity pairs: $(R_B, \bm{\omega}_B)$, $(R_{A_L}, \bm{\omega}_{A_L}^{B})$, $(R_{A_R}, \bm{\omega}_{A_R}^{B})$, $(R_{W_L}, \bm{\omega}_{W_L}^{A_L})$, and $(R_{W_R}, \bm{\omega}_{W_R}^{A_R})$. 

Let $\theta$ angles be the biologically meaningful flapping angles, as illustrated in Fig. \ref{fig:aerobat_dof}, and the superscript $L$ and $R$ represents the left and right wing joint angles respectively. Then the left armwing rotation matrices are defined as follows:
\begin{equation}
    \begin{aligned}
        R_{A_L} &= R_z(\theta^L_m)\,R_x(\theta^L_p),\, &
        R_{W_L} &= R_x(\theta^L_e)\,R_z(\theta^L_f),
    \end{aligned}
\label{eq:rotation_matrices_left}
\end{equation}
where $R_x(\theta)$ and $R_z(\theta)$ are the rotational matrix about $x$ and $z$ axis respectively. The left armwing angular velocities are defined as follows:
\begin{equation}
    \begin{aligned}
        \bm{\omega}_{A_L}^{B} &= \begin{bmatrix} 0, 0, \dot{\theta}_m^L \end{bmatrix}^\top + R_z(\theta^L_m)  \begin{bmatrix} \dot{\theta}_p^L, 0 , 0 \end{bmatrix}^\top + \bm{\omega}_{B}^{B}\\
        \bm{\omega}_{W_L}^{A_L} &= \begin{bmatrix} 
        \dot{\theta}_e^L, 0 , 0 \end{bmatrix}^\top + R_x(\theta^L_e) 
        \begin{bmatrix} 0, 0, \dot{\theta}_f^L \end{bmatrix}^\top + \bm{\omega}_{A_L}^{A_L}.
    \end{aligned}
\label{eq:angular_vel_left}
\end{equation}
The right wing derivations can be derived in a similar fashion. Therefore, for the rest of the paper, only the left-wing components will be derived if the right side also follows a similar derivation.

As shown in Fig. \ref{fig:aerobat_model}, let $\bm p_B$ be the linear position of the center of mass of a body, $\bm{l}_{Lj}$ and $\bm{l}_{Rj}$, $j = \{1,2,3\}$, be the length vectors that represent the morphology of the Aerobat mechanism, which are constant with respect to their local frame of reference. Then the linear position of the center of mass of the left armwing can be derived as follows:
\begin{equation}
    \begin{aligned}
        \bm{p}_{A_L} &= \bm{p}_B + R_{B}\,\bm{l}_{L_1}^{B} + \tfrac{1}{2}\,R_B\,R_{A_L}\,\bm{l}_{L_2}^{A_L} \\
        \bm{p}_{W_L} &= \bm{p}_{A_L} + \tfrac{1}{2}\,R_B\,R_{A_L}\,\bm{l}_{L_2}^{A_L} + R_B\,R_{A_L}\,R_{W_L}\,\bm{l}_{L_3}^{W_L},
    \end{aligned}
\label{eq:linear_position}
\end{equation}
The linear velocity of the center of mass can be derived from \eqref{eq:linear_position} by differentiating the linear positions with respect to time. Note that the linear positions and velocities are defined with respect to the inertial frame. 


The kinetic and potential energy of the system can be derived as follows.
\begin{equation}
    \begin{aligned}
    T &= \sum_{F \in \mathcal{F}} \left( m_F\,\dot{\bm{p}}_F^\top\,\dot{\bm{p}}_F + (\bm{\omega}_F^{F})^\top\,\hat{I}_{F}\,\bm{\omega}_F^{F} \right)\frac{1}{2} \\
    U &= \sum_{F \in \mathcal{F}} m_F \, [ 0, 0, g ] \, \bm{p}_F,
    \end{aligned}
\label{eq:energy}
\end{equation}
where $\mathcal{F} = \{B, A_L, A_R. W_L, W_R\}$ is the set containing the frame of references, $m_F$ and $\hat{I}_{F}$ are the mass and inertia matrix of the corresponding body respectively. $\hat{I}_{F}$ is defined about the local frame of reference which is diagonal and constant. Then the Lagrangian of the system, $L = T - U$, can be used to derive the equation of motion. 

The body rotation $(R_B,\omega_B)$ is derived using the modified Euler-Lagrangian formulation for a rotation in SO(3). This formulation is not susceptible to gimbal lock which might happen if we use Tait-Bryan angles during the upside-down maneuver. The modified Euler-Lagrange equation for rotation in SO(3) can be derived by using Hamilton's principle \cite{lee2017global}, which has the following form:
\begin{equation}
\begin{gathered}
    \frac{d}{dt}\frac{\partial L}{\partial \bm{\omega}_B^B} + \bm{\omega}_B^B \times \frac{\partial L}{\partial \bm{\omega}_B^B} + \sum^3_{j=1}\bm{r}_{B,j} \times \frac{\partial L}{\partial \bm{r}_{Bj}} = \bm{\tau}_B^B \\
    \dot{R}_B = R_B\,S(\bm{\omega}_B^B),
\end{gathered}
\label{eq:eom_hamiltonian}
\end{equation}
where $S(\cdot)$ is a skew operator, $R_B^\top = [\bm{r}_{B1}, \bm{r}_{B2}, \bm{r}_{B3}]$ and $\bm{\tau}_B^B$ is the non-conservative torque about the generalized coordinate $\bm{\omega}_B^B$. The equation of motion of the remaining states can be solved by using the Euler-Lagrange equation:
\begin{equation}
    \begin{gathered}
        \bm{\theta}_L = [\theta_p^L, \theta_m^L, \theta_e^L, \theta_f^L], \qquad \bm{\theta}_R = [\theta_p^R, \theta_m^R, \theta_e^R, \theta_f^R], \\
        \bm{q}_{e} = [\bm{p}_B^\top, \bm{\theta}_L^\top, \bm{\theta}_R^\top]^\top, \qquad
        \frac{d}{dt}\frac{\partial L}{\partial \dot{\bm{q}}_{e}} - \frac{\partial L}{\partial \bm{q}_{e}} = \bm{u}_{e},
    \end{gathered}
\label{eq:eom_euler_lagrangian}
\end{equation}
where $\bm{u}_{e}$ is the non-conservative force about the generalized coordinate $\bm{q}_{e}$. Combining \eqref{eq:eom_hamiltonian} and \eqref{eq:eom_euler_lagrangian}, the equation of motion can be formulated into the following form:
\begin{equation}
    \begin{gathered}
    \bm{q} = [\bm{r}_B^\top, \bm{p}_B^\top, \bm{\theta}_L^\top, \bm{\theta}_R^\top]^\top, \qquad
    \bm{q}_d = [ \bm{\omega}_{B}^\top, \dot{\bm{p}}_B^\top, \dot{\bm{\theta}}_L^\top, \dot{\bm{\theta}}_R^\top]^\top \\
    M\,\dot{\bm{q}}_d + C\bm{q}_d + G = B_a\, \bm{u}_a + B_m\, \bm{u}_m,
    \end{gathered}
\label{eq:eom_summary}
\end{equation}
where $M$ and $C$ denote the mass-inertial and Coriolis matrices. $G$ is the gravity vector and $\bm{r}_B$ is the rotation matrix $R_B$ concatenated into a vector form. In Eq.~\ref{eq:eom_summary},  
\begin{itemize}
    \item $\bm{u}_a$ is the generalized aerodynamic forces and torque. 
    \item $\bm{u}_m$ is the generalized motor torque acting on the armwing joints which is selected to directly actuate the joints angles $\bm \theta_L$ and $\bm \theta_R$.
\end{itemize}
Specifically speaking, we aim to estimate $\bm{u}_a$. $\bm{u}_m$ embodies constrained joint torques, that is,  
\begin{equation}
    \begin{aligned}
    \bm{u}_m &= [\tau_p^L, \tau_m^L, \tau_e^L, \tau_f^L, \tau_p^R, \tau_m^R, \tau_e^R, \tau_f^R]^\top \\
    B_m &= [0_{8 \times 6}, I_{8 \times 8}]^\top,
    \end{aligned}
\end{equation}
where $\tau$ represents the torque that acts on each joint. Since joint movements are known (wing kinematics using joint encoders), joint torques can be computed. 

Next, we design two estimators based on conjugate momentum observers and the regression method to estimate aerodynamic forces in Eq.~\ref{eq:eom_summary}.

\section{External Force Estimation}
\label{sec:estim}

\subsection{Conjugate Momentum Method}

The general idea behind conjugate momentum observers is that the signal $r(t)$ can be computed based on vehicle states to carry information about external aerodynamic forces on the vehicle.

Let's revisit the equations of motion given by
\begin{equation}
M\,\dot{\bm{q}}_d + C\bm{q}_d + G = B_a\, \bm{u}_a + B_m\, \bm{u}_m,
\end{equation}
We construct the signal $r(t)$
\small
\begin{equation}
r(t) = K \left( p(t) - \int_0^t \left(\bm{u}_m + C^T(\bm{q},\bm{q}_d)\bm{q}_d - G(\bm{q}) + r(s)\right) ds - p(0) \right)
\end{equation}
where $K$ is the gain matrix (fixed diagonal matrix). $p(t)$ denotes conjugate momentum variable and is computed by multiplying the mass-inertia matrix ($M$) and the generalized coordinate velocity ($\bm{q}_d$).
\begin{figure}[t]
  \centering
  \includegraphics[width=1.0\columnwidth]{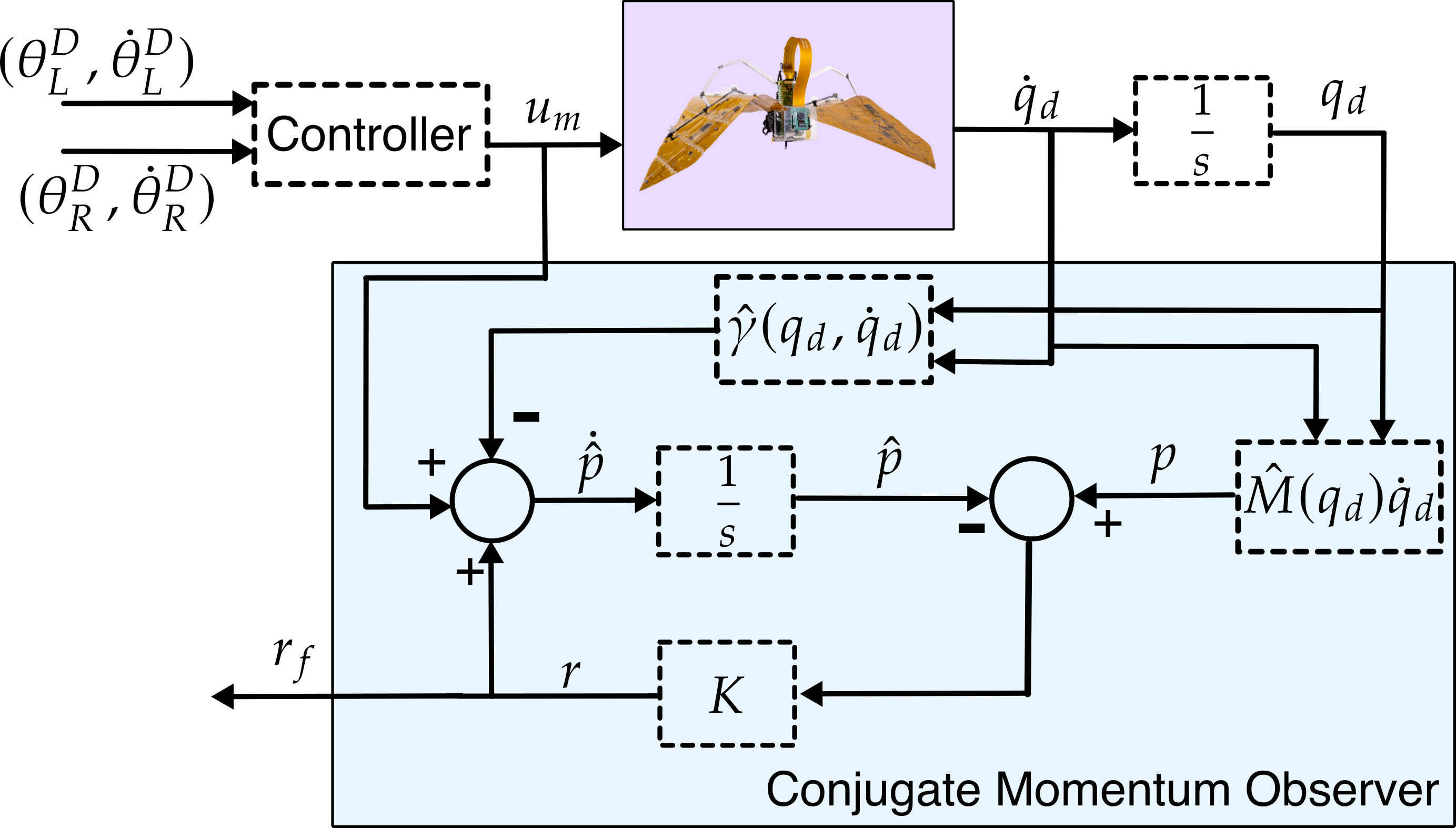}
  \caption{Block diagram of the conjugate momentum-based observer architecture for the Aerobat. The controller receives desired joint trajectories $(\theta^D_L, \dot{\theta}^D_L)$ and $(\theta^D_R, \dot{\theta}^D_R)$ and generates motor torque $u_m$. The observer (blue block) estimates the momentum $\hat{p}$ using the nominal dynamics $\hat{M}(q_d)\dot{q}_d$ and Coriolis/centrifugal terms $\hat{\gamma}(q_d, \dot{q}_d)$. The error in momentum is fed through a gain $K$ to produce a residual signal $r$, which contributes to the feedback term $r_f$.}

  \label{fig:block}
\end{figure}
Now, it can be shown that $r(t)$ is a low-pass-filtered version of the aerodynamic force $\bm{u}_a$, that is,
\begin{equation}
\dot{r} = Kr + K\bm{u}_a 
\end{equation}
To prove this relationship, we use the following known equation.
\begin{equation}
\dot{M} = C(\bm{q},\bm{q}_d) + C^T(\bm{q},\bm{q}_d)
\end{equation}
and the definition of conjugate momentum $p = M(\bm{q})\bm{q}_d$.

\subsection{Observer Stability Analysis}

Assume that the true residual is 
\[
r = B_a \bm{u_a},
\]
and we design an observer for this residual as 
\[
\dot{\hat{r}} = K_0 \Bigl( B_a \bm{u_a} - \hat{r} \Bigr),
\]
where \(K_0\) is a positive definite gain matrix. Define the estimation error as 
\[
e = B_a \bm{u_a} - \hat{r}.
\]
Taking the time derivative of the error yields
\[
\dot{e} = \frac{d}{dt}\bigl( B_a \bm{u_a} \bigr) - \dot{\hat{r}}.
\]
Assuming that the rate of changes in \(B_a \bm{u_a}\) is smaller compared to the observer dynamics (two- time scale dynamics), we have
\[
\dot{e} = 0 - K_0 \Bigl( B_a \bm{u_a} - \hat{r} \Bigr) = -K_0 e.
\]
The differential equation becomes very fast in the dynamics of $e$ if $K_0$ is very large and its solution is given by  
\[
\dot{e} = -K_0 e
\]
\[
e(t) = e(0) \exp(-K_0 t).
\]
Since \(K_0\) is positive definite, the error \(e(t)\) decays exponentially to zero, ensuring that \(\hat{r}\) converges to \(\bm{u_a}\). Thus, the observer successfully tracks the aerodynamic force input. 

\subsection{Regression Method}
The force estimation model is obtained using an MLP to approximate the mapping of actuation and experimental parameters to measured aerodynamic forces. The loss function is the mean squared error (MSE), defined as:

\begin{equation} 
\mathcal{L}(\Theta) = \frac{1}{N} \sum_{t=1}^{N} \left( \hat{\mathbf{y}}_t - \mathbf{y}_t \right)^2 
\end{equation}

where $N$ is the total number of data points in the training dataset, $\mathbf{y}_t \in \mathbb{R}^3$ represents the ground truth force at the time step $t$, $\hat{\mathbf{y}}_t$ is the predicted force, and $\Theta$ denotes the network parameters trained via gradient descent.

The input to the MLP is a feature vector $\mathbf{X}_t \in \mathbb{R}^{5}$, where the five inputs correspond to the joint angles $\mathbf{\theta}_{L}(t)$ and $\mathbf{\theta}_{R}(t)$, flapping frequency, pitch angle, and the wind speed. The inputs are normalized before being passed into the model. The model output is a force vector $\hat{\mathbf{y}}_t = [\hat{F}_x, \hat{F}_y, \hat{F}_z]^\top$.

The MLP consists of two hidden layers, each containing 64 neurons with SiLU activation. The inference results are presented in Section \ref{sec:res}.

\section{Results}
\label{sec:res}



\begin{figure}[t]
  \centering
  \includegraphics[width=1.0\columnwidth]{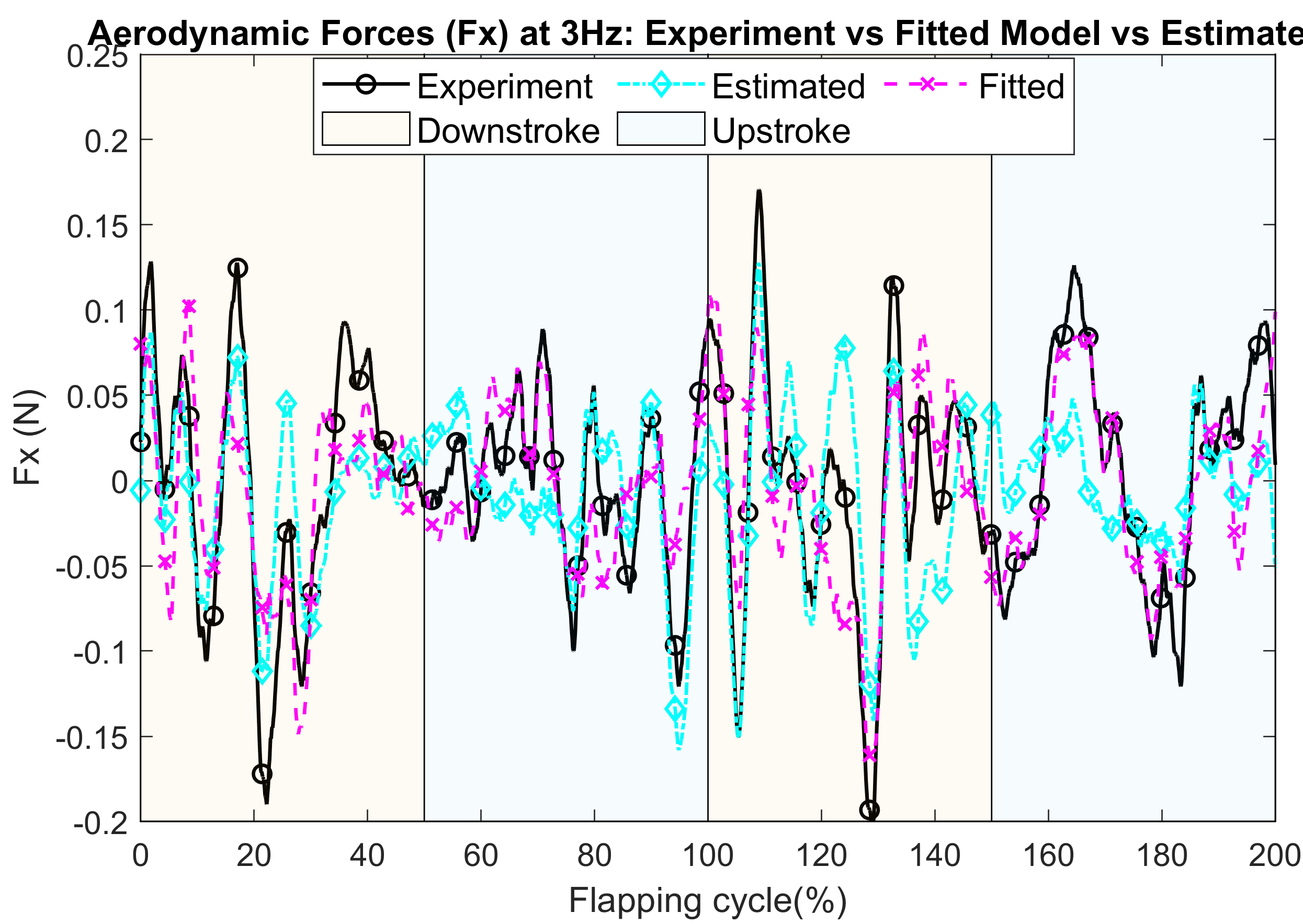}
  \caption{ Evaluation of aerodynamic force estimation ($F_x$) using the conjugate momentum method and an MLP-based regression model against experimental data.}
  \label{fig:Fx_NN_Conjg}
\end{figure}

\begin{figure}[t]
  \centering
  \includegraphics[width=1.0\columnwidth]{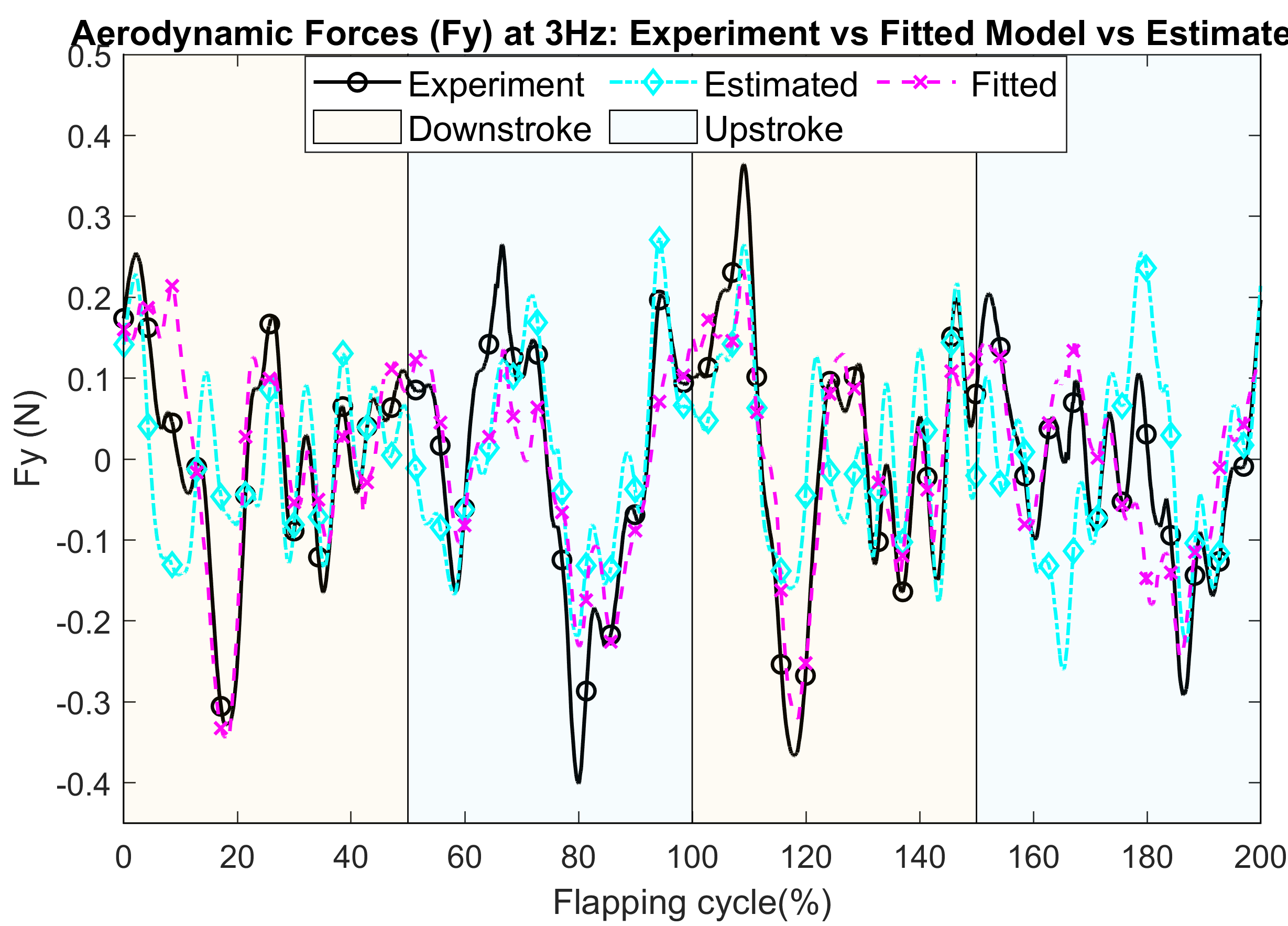}
  \caption{ Evaluation of aerodynamic force estimation ($F_y$) using the conjugate momentum method and an MLP-based regression model against experimental data.}
  \label{fig:Fy_NN_Conjg}
\end{figure}

\begin{figure}[t]
  \centering
  \includegraphics[width=1.0\columnwidth]{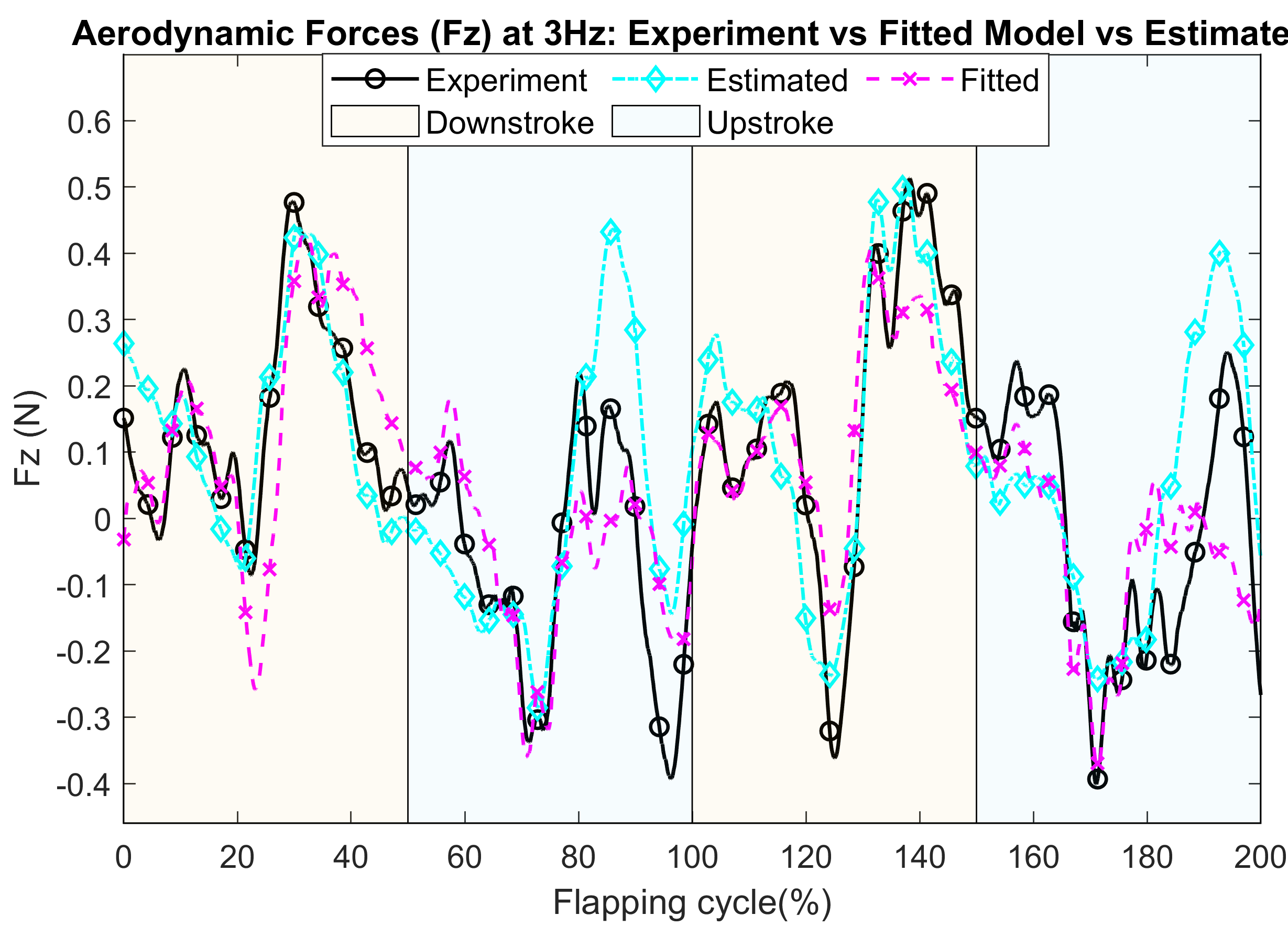}
  \caption{ Evaluation of aerodynamic force estimation ($F_z$) using the conjugate momentum method and an MLP-based regression model against experimental data.}
  \label{fig:Fz_NN_Conjg}
\end{figure}


Using the load cell setup illustrated in Fig.~\ref{fig:loadcell}, we performed experiments with Aerobat flapping at a frequency of 3~Hz. In these tests, Aerobat was rigidly mounted to the load cell to maintain a constant position and orientation. An encoder measured the absolute rotation of the actuator shaft, which were converted into joint angles via established kinematic relationships. Force data were sampled at 7000~Hz over five consecutive experiments. In addition, a broader dataset was collected under varied operating conditions, including flapping frequencies of 2.5~Hz, 3~Hz, 3.5~Hz, and 4~Hz; wind speeds of 0.5~m/s, 1~m/s, 1.5~m/s, and 2~m/s; and robot pitch angles of $0^\circ$, $-5^\circ$, $-10^\circ$, $-15^\circ$, and $-20^\circ$, totaling 40 datasets.
The dataset was split by condition, with 80\% of the experimental conditions used for training the MLP-based regression model and the remaining 20\% reserved for testing. The same dataset was employed to validate the conjugate momentum observer.The regression model uses a multi-layer perceptron (MLP) with three hidden layers of size $128 \times 128 \times 128$, trained using mean squared error (MSE) loss and the Adam optimizer. The hyperparameters were tuned empirically based on validation performance.

The mass-inertia matrix, Coriolis terms, and gravity vector were computed and integrated into the estimator model, which is based on the conjugate momentum observer detailed in Section~\ref{sec:estim}.In our implementation (shown in Fig.~\ref{fig:block}), the desired joint angles and their corresponding velocities are provided to a dedicated controller that computes the motor torque commands, $u_m$, for the Aerobat model. This observer then computes a filtered version of the aerodynamic force signal by exploiting the relationship between the conjugate momentum and the external aerodynamic loads, enabling more accurate force estimation for real-time control and optimization.

Figures ~\ref{fig:Fx_NN_Conjg}, ~\ref{fig:Fy_NN_Conjg} and ~\ref{fig:Fz_NN_Conjg} presents a detailed comparison of the aerodynamic force components $F_x$, $F_y$, and $F_z$ measured experimentally (denoted by black markers), predicted by a conjugate momentum--based estimator (cyan markers) and fitted using a regression model (magenta markers). The forces are plotted over two consecutive flapping cycles (0\%--200\% of the wingbeat), with shaded regions clearly distinguishing the downstroke and upstroke phases. This comprehensive visualization underscores the periodic nature of the aerodynamic forces and provides insight into how each estimation method captures the dynamic force profile. The experimental data serve as a benchmark, while the estimators are evaluated for their ability to replicate the observed force peaks, transitions, and overall trends. 

Across the three force directions, the measured aerodynamic forces exhibit pronounced periodicity, reflecting the cyclic nature of flapping flight. During the downstroke (highlighted in a lighter shade), forces typically reach higher magnitudes---particularly evident in $F_z$ due to the increased interaction of wing surface with the surrounding air. In Contrast, during the upstroke (darker shade), negative or lower forces frequently appear, indicative of reduced aerodynamic lift and possible inertial effects arising from rapid wing retraction. The physics-based estimator (cyan) generally reproduces these cyclical patterns, albeit with an occasional underestimation of peak forces. Such discrepancies may arise from unmodeled fluid-structure interactions, rigid-body approximations, or uncertainties in wing kinematics. Meanwhile, the data-driven model (magenta) often captures abrupt changes and peak values more accurately, particularly around transitions between downstroke and upstroke, owing to its capacity to learn nuanced, non-linear relationships directly from the experimental dataset.

Overall, the close alignment of the physics-based and data-driven estimations with the experimental measurements underscores the viability of these methods for real-time force prediction in flapping-wing systems. Quantitatively, the regression model achieved RMSE values of $0.0401$, $0.0696$, and $0.1155$ for $F_x$, $F_y$, and $F_z$, respectively, while the conjugate momentum--based estimator recorded RMSE values of $0.0467$ for $F_x$, $0.1059$ for $F_y$, and $0.1322$ for $F_z$. These metrics indicate that the data-driven approach provides slightly higher precision, particularly in capturing the complex dynamics of $F_y$ and $F_z$. However, the conjugate momentum method, with its inherent physical interpretability and lower dependency on extensive training data, remains a robust alternative for flight scenarios. Minor deviations, such as phase shifts and amplitude mismatches, reflect the inherent challenges in modeling unsteady aerodynamics, wing flexibility, and rapid inertial changes. Together, these results demonstrate that both estimation frameworks can robustly replicate the force profiles generated by a bio-inspired, dynamically morphing wing, providing valuable insights for subsequent control, optimization, and design efforts in real flight scenarios.

\section{Concluding Remarks}

In this work, we presented and validated two complementary approaches for estimating aerodynamic forces in dynamically morphing wing flight using the Aerobat platform: a physics-based conjugate momentum observer and a data-driven neural network regression model. Accurate force estimation remains a critical step in enabling agile control and design optimization of flapping-wing robots, particularly those with bio-inspired morphing geometries that experience complex unsteady fluid-structure interactions.

The conjugate momentum observer, grounded in Hamiltonian mechanics, leverages system dynamics and sensor feedback to reconstruct external force input without relying on extensive datasets. This method demonstrated robustness in multiple trials and offered physically interpretable estimates that align well with measured ground-truth forces. Although the approach exhibited slight deviations in peak force estimation--primarily in Fy and Fz--its consistent performance and independence from data-intensive training make it attractive for real-time onboard implementation.

On the other hand, the neural network–based regression method, trained on a diverse dataset incorporating flapping frequency, joint kinematics, and environmental variables, demonstrated superior accuracy in capturing rapid transients and nonlinear patterns in the aerodynamic force profiles. By achieving lower RMSE values in all force components, the regression model highlights the potential of learning-based estimators to improve predictive performance, especially in highly dynamic and underactuated aerial systems.

Experimental results confirm that both frameworks are capable of reproducing the cyclical nature of flapping flight and the directional asymmetries arising from morphing wing motion. These findings establish a solid foundation for future work in force-aware control, energy-efficient gait optimization, and autonomous flight planning in morphing-wing aerial robots.

Looking ahead, we plan to integrate these estimators with onboard control algorithms for untethered flight experiments and extend the modeling framework to account for fluid-structure coupling effects and flexible-body dynamics. Moreover, hybrid estimation strategies that fuse physical priors with learning-based corrections hold promise in balancing generalization with interpretability. Ultimately, this work contributes to a validated set of tools to advance closed-loop autonomy in insect- and avian-inspired aerial systems.

\section{ACKNOWLEDGMENT}
This work is supported in part by the U.S. National Science Foundation (NSF) under CAREER Award No. 2340278 and Grants No. CMMI-2142519 and CMMI-2140650.


\printbibliography

\end{document}